%% file: emnlp2021.tex
\pdfoutput=1

\documentclass[11pt]{article}

\usepackage[]{emnlp2021}

\usepackage{times}
\usepackage{latexsym}

\usepackage[T1]{fontenc}

\usepackage[utf8]{inputenc}

\usepackage{microtype}

\usepackage{graphicx}
\usepackage{url}
\usepackage{amsmath}

\usepackage{booktabs}
\usepackage{multirow}
\usepackage{enumitem}

\usepackage{amssymb}

\title{Efficient Contrastive Learning via Novel Data Augmentation and Curriculum Learning}

\author{Seonghyeon Ye, Jiseon Kim, Alice Oh \\
 School of Computing, KAIST \\
 \texttt{\{vano1205, jiseon\_kim\}@kaist.ac.kr} \\
 \texttt{alice.oh@kaist.edu}
}

\begin{document}
\maketitle
\begin{abstract}
\input{abstract}
\end{abstract}

\section{Introduction}
\input{introduction.tex}


\section{Method} 
\input{method.tex}

\section{Experiments}
\input{experiments.tex}

\section{Results}
\input{results.tex}

\section{Conclusion}
\input{conclusion.tex}

\section*{Acknowledgements}
The authors want to thank Kyubeom Han for helpful discussions and support. This work was supported by Institute of Information \& communications Technology Planning \& Evaluation (IITP) grant funded by the Korea government(MSIT) 
(No.2017-0-01779, A machine learning and statistical inference framework for explainable artificial intelligence).
\bibliography{anthology,custom}
\bibliographystyle{acl_natbib}

\clearpage
\appendix
\section{Appendix}
\label{sec:appendix}
\input{appendix}


\end{document}

%% file: abstract.tex
We introduce EfficientCL, a memory-efficient continual pretraining method that applies contrastive learning with novel data augmentation and curriculum learning. For data augmentation, we stack two types of operation sequentially: cutoff and PCA jittering. While pretraining steps proceed, we apply curriculum learning by incrementing the augmentation degree for each difficulty step. After data augmentation, we apply contrastive learning on projected embeddings of original and augmented examples. When fine-tuned on GLUE benchmark, our model outperforms baseline models, especially for sentence-level tasks. Additionally, this improvement is achieved with only 70\% of computational memory compared to the baseline model. \footnote{Our code is publicly available at \url{https://github.com/vano1205/EfficientCL}}

%% file: introduction.tex
Many state-of-the-art language models involve the paradigm of unsupervised pretraining followed by fine-tuning on downstream tasks \cite{BERT, liu2019roberta, GPT3}. However, pretraining a language model from scratch with a huge corpus has high computational costs. One way to cut down the cost is to use continual training of a pretrained model which could improve the language model with less computation \cite{DeCLUTR}.

Contrastive learning is effective for self-supervised learning for image classification \cite{SimCLR, SimCLRv2}, and it works by allowing the model to put similar examples close and different examples far from one another. Often in contrastive learning, data augmentation is used to make the positive pairs. Recent papers describe how to apply contrastive learning to the language domain \cite{COCO-LM, SupervisedCL, CoDA, CLEAR}, and even a combination of contrastive learning with continual pretraining \cite{DeCLUTR}. However, because of the sequential nature of language, it is difficult to apply data augmentation methods used in images directly to language modeling. 

Additionally, curriculum learning is a powerful training technique for deep networks \cite{CurrPower, CAT}. By training easy to hard examples in order, it facilitates faster convergence, leading to better performance. Other studies show that it is also effective for language modeling \cite{NLPCurr, EDA, Shortformer, GPTCurr}. 

We propose an efficient yet powerful continual pretraining method using contrastive learning and curriculum learning. The contribution of this paper is as follows:
\begin{itemize}[noitemsep,nolistsep]
    \item We suggest a novel data augmentation method for contrastive learning: first cutoff, then PCA jittering.
    This leads to robustness to sentence-level noise, resulting in a better sentence-level representation.
    \item We apply curriculum learning by increasing the noise degree of augmentation for each level of difficulty. This leads to faster convergence at the pretraining stage.
    \item In addition to outperforming baseline models on the GLUE benchmark, our model is memory efficient and applicable to a wider range of corpora. 
\end{itemize}

%% file: method.tex
\begin{figure*}[!ht]
\centering
\includegraphics[width=0.8\textwidth]{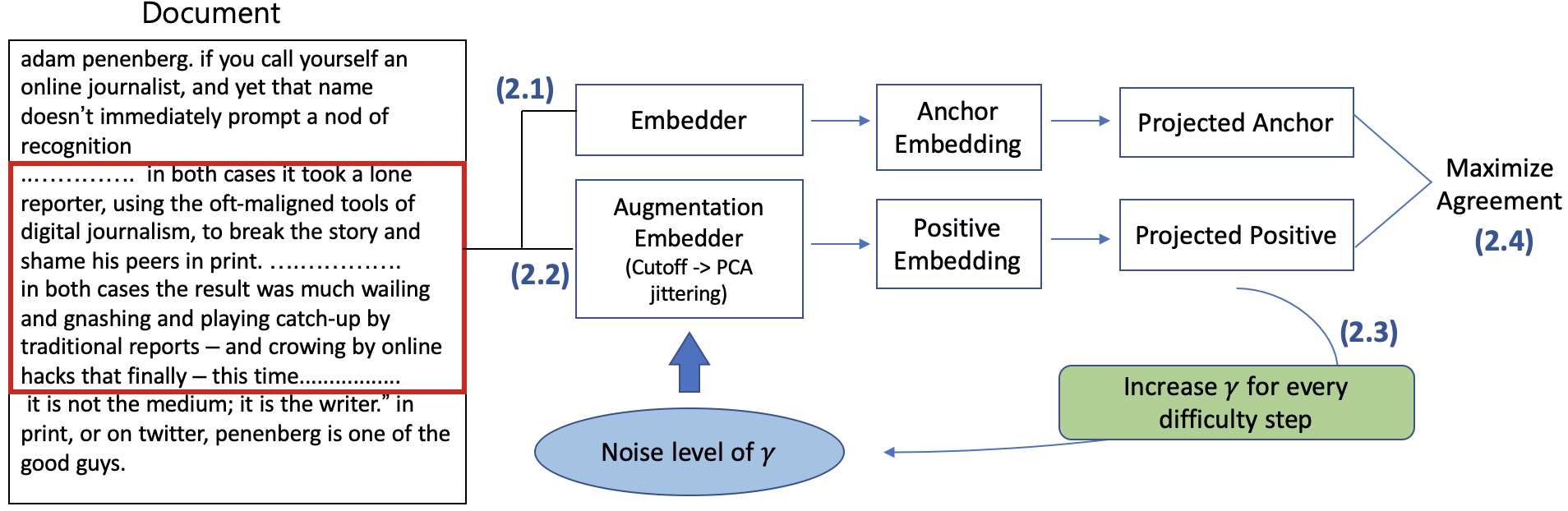}
\caption{For continual pretraining, we sample a fixed-length sequence from each document and obtain the anchor embedding. We obtain the positive embedding from the augmentation embedder with cutoff and PCA jittering. While training, we increase the augmentation degree for every difficulty step. From projected anchor and positive embedding, we apply contrastive learning to maximize agreement.}
\label{fig:model}
\vspace{-5mm}
\end{figure*}

The overall learning process of EfficientCL is illustrated in Figure \ref{fig:model}.

\subsection{Sampling Anchor from Text}
\label{sampleanchor}
We modify the method from \citet{DeCLUTR} which samples anchors and positive instances. For each document from the pretraining corpus, a sequence of 512 tokens is randomly sampled, referred to as an anchor. The anchor goes through the RoBERTa encoder to get the embedding of the anchor sequence.

\subsection{Data Augmentation}
\label{dataaugmentation}
We introduce two data augmentation methods, \textit{cutoff} and \textit{PCA jittering}. These methods were inspired by SimCLR  \cite{SimCLR}, which showed that random cropping followed by color jittering is effective for contrastive learning among various augmentation combinations. Likewise, our method applies cutoff first, then PCA jittering. We apply on one of the inner layers of RoBERTa model randomly sampled from \{7,9,12\} layers. These layers contain the most of syntactic and semantic information \cite{Mixtext}. 

\paragraph{\noindent\textbf{Cutoff Augmentation}}
Cropping-based data augmentation is simple but effective for natural language understanding \cite{COCO-LM, Cutoff}. Among various methods introduced in \citet{Cutoff}, we use the most robust \textit{span cutoff} method. Previous span cutoff method makes a continuous portion of sequences with a specific ratio to zero on the embedding layer. On the other hand, we apply this operation to the hidden states of an inner layer of RoBERTa. 
\vspace{-2mm}
\paragraph{\noindent\textbf{PCA jittering Augmentation}} 
Our method is similar to the widely-used color jittering method in \citet{AlexNet} for computer vision domain, but we apply the operation on the hidden states.
If the original hidden state is \(h=[h_0, h_1, ..., h_d]\), hidden state after PCA jittering would be \(h=[h_0+ \delta, h_1+\delta, ..., h_d+\delta]\) 
where \(\delta = [p_1, p_2, ..., p_d][\alpha\lambda_1,\alpha \lambda_2, ..., \alpha\lambda_d]^T\), \(\alpha \sim N(0, \sigma^2)\), \(d\) is the dimension of hidden states, \(p_i\) and \(\lambda_i\) are the \(i\)th eigenvector and eigenvalue respectively.

\begin{table*}[!t]
\small
\centering
\begin{tabular}{@{}l|rrrrrrrr|r@{}}
\toprule
Model & CoLA & MNLI & MRPC & QNLI & QQP & RTE & SST & STSB & Avg. \\ \midrule
\textbf{Roberta-base} & 59.38 & 87.76 & 90.07 & 92.44 & 89.44 & 76.90 & 94.15 & 90.55 & 85.09 \\
\textbf{DeCLUTR} & \textbf{59.60} & 87.75 & 89.63 & 92.70 & \textbf{89.58} & 75.63 & 94.04 & 90.48 & 84.93 \\ 
\textbf{EfficientCL} & 59.05 & \textbf{87.80} & \textbf{90.68} & \textbf{92.81} & 89.51 & \textbf{77.62} & \textbf{94.61} & \textbf{90.61} & \textbf{85.34} \\ 
\bottomrule
\end{tabular}
\caption{Evaluation result on GLUE development set. All performance is median of five runs with random seeds except RTE, which reports median of ten runs. The evaluation metric for each task is as follows: Matthews correlation for CoLA, average of Pearson and Spearman correlation for STS, average of accuracy and F1 for MRPC and QQP, and accuracy for other tasks.}
\label{table:overall_performance}
\vspace{-5mm}
\end{table*}

\subsection{Simple Curriculum Learning Method}
\label{curriculum}
We apply curriculum learning during the data augmentation process by increasing the noise level for each difficulty step, which is cropping ratio for cutoff and standard deviation of the noise hyperparameter for PCA jittering method. As the noise level gets larger, the augmented positive would be more dissimilar from the anchor, resulting in a harder example for contrastive learning. For both augmentation methods, the noise level is initially set as 0.01 and incremented until 0.1.
\footnote{The rationale is explained in Appendix \ref{sec:hyperparam}.}

We introduce and compare two curriculum learning methods: \textit{Discrete} and \textit{Continuous}. First, the discrete curriculum learning divides the pretraining step into ten steps and increases the augmentation level for each step.
Second, the continuous curriculum learning increases the augmentation level continuously for every iteration of training, starting from 0.01 until 0.1. 

\subsection{Contrastive Learning Framework}
\label{contrastive}
For each positive and anchor sequence embedding denoted as \(e_i\), the projected embedding is obtained by \(z_i=g(e_i)=W^{(2)}\psi(W^{(1)}e_i)\) where \(W^{(1)},W^{(2)}\) are trainable weights and \(\psi\) is a ReLU non-linearity \cite{SimCLR}.
From the projected anchor and the projected positive, we apply contrastive learning. For a minibatch of size \(N\), there is an anchor and a positive instance for each document, resulting in 2\(N\) instances in total. 
The contrastive loss for a positive pair \(\{z_{2i-1},z_{2i}\}\) is 
\[L_{contrastive}=\sum_{i=1}^{N}{l(2i-1,2i)+l(2i,2i-1)}\]
\[l(i,j)=-log\frac{\mathrm{exp}(\mathrm{sim}(z_i,z_j)/\tau)}{\sum_{k=1}^{2N}1_{[i\neq k]}\mathrm{exp}(\mathrm{sim}(z_i,z_k)/\tau)}\]
\[\mathrm{sim}(z_i,z_j) = z_i^Tz_j/(\left\|z_i\right\|\left\|z_j\right\|)\]
where \(\tau\) refers to the temperature hyperparameter. 

Our training process is applied continuously on the pretrained RoBERTa model. To prevent catastrophic forgetting of token-level MLM objective \cite{Forgetting},
we add the loss from the MLM objective to the contrastive objective, 
\[L_{total}=L_{MLM}+L_{contrastive}.\]

%% file: experiments.tex
\subsection{Pretraining and Finetuning}
The dataset used for pretraining is OpenWebText corpus containing 495,243 documents with a minimum token length of 2048, which is the same setting as the DeCLUTR model \cite{DeCLUTR}. 
We use a NVIDIA Tesla V100 GPU for pretraining, which takes 19.7 hours for training. For finetuning evaluation, we use the development set in GLUE benchmark. For small datasets (CoLA, STSB, MRPC, RTE), the model is finetuned for 10 epochs, and for the rest (MNLI, QQP, SST, QNLI), it is trained for 3 epochs. 
We report median values over 5 random initializations for all tasks except RTE. For RTE, we report median values of 10 runs due to the high variance of the performance.  

\subsection{Baselines}

We compare our model with the pretrained RoBERTa-base to check the effectiveness of our continual pretraining method. We observe whether our contrastive objective complements the MLM objective by improving the performance of sentence-level tasks. We also compare with the DeCLUTR model \cite{DeCLUTR} which is continually pretrained from RoBERTa-base with contrastive learning. 
It samples adjacent positive sequences given an anchor instead of applying data augmentation or curriculum learning. 
\footnote{Additional variants of baseline models are shown in Appendix \ref{sec:variant_baseline}.}

%% file: results.tex
\subsection{Overall Results}
We compare the performance of our model with two baseline models, and the results are shown in Table \ref{table:overall_performance}. Overall, our model performs better than the baseline models, especially for small datasets such as MRPC and RTE. For sentence-level tasks (all except CoLA), our model performs better than RoBERTa-base and better than DeCLUTR except QQP. Since our pretraining method makes the model robust to sentence-level noise, it captures better representations of sentences. 
For the CoLA dataset, our model performs poorly because it is trained to be robust from small to big noises sequentially. The linguistic acceptability task is sensitive to noise, meaning that small changes in a sentence could lead to a different label. Our method hardly differentiates when the change is small, leading to wrong predictions.

\subsection{Ablation Study}
We conduct ablation studies on the GLUE development set, and the results are shown in Table \ref{table:ablation_augmentation}.

\begin{table*}[!t]
\small
\centering
\begin{tabular}{@{}cccc|rrrrrrrr|r@{}}
\toprule
Curriculum & Cutoff & PCA & EDA & CoLA & MNLI & MRPC & QNLI & QQP & RTE & SST & STSB & Avg. \\ \midrule
Discrete & \checkmark & & & 58.04 & 87.69 & 90.07 & 92.60 & 89.54 & 76.90 & 94.27 & 90.42 &  84.94\\
Discrete & &\checkmark & & 59.07 & 87.56 & 89.89 & 92.62 & 89.44 & 76.17 & 94.15 & 90.48 & 84.92 \\ 
Discrete & & & \checkmark & 58.04 & 87.72 & 90.27 & 92.66 & 89.49 & 76.71 & 93.92 & 90.49 & 84.91 \\
No Curr &\checkmark & \checkmark & & \textbf{60.33} & 87.70 & 90.27 & 92.51 & 89.51 & 76.17 & 94.04 & 90.51 & 85.13 \\
Continuous &\checkmark & \checkmark & & 58.82 & 87.71 & 90.27 & 92.60 & \textbf{89.57} & 76.35 & 94.27 & 90.45 & 85.01\\ 
Discrete& \checkmark & \checkmark& & 59.05 & \textbf{87.80} & \textbf{90.68} & \textbf{92.81} & 89.51 & \textbf{77.62} & \textbf{94.61} & \textbf{90.61} & \textbf{85.34}\\ 
\bottomrule
\end{tabular}
\caption{Ablation studies on GLUE dev set. The first three rows show the results of different data augmentation. The fourth and fifth test the impact of curriculum learning. The last setting is the best, using both cutoff and PCA jittering and discrete curriculum learning.}
\label{table:ablation_augmentation}
\vspace{-5mm}
\end{table*}
\paragraph{\textbf{Data Augmentation Method}}
To observe the impact of each augmentation method on the performance, we conduct experiments with only one of the two methods. For all tasks except QQP, each method underperforms the combination of both. This is consistent with \citet{SimCLR} which suggests that a composition of multiple augmentations is effective. Because each of the augmentations is relatively simple, applying only one method leads to a small degree of augmentation. This disturbs effective learning because the benefit of \textit{hard positive} in contrastive learning is neglected.

We highlight the novelty of our augmentation method by comparing with \citet{EDA} which uses the popular EDA augmentation method. Our method is different from \citet{EDA} in that curriculum learning and data augmentation are applied at the continual pretraining stage, not at the finetuning stage. The augmentation level is from 0 to 0.5 with 6 discrete steps, the same as the paper. The results show that EDA underperforms our method for all tasks, and although EDA is a simple data augmentation approach, it does not perform well in the continual pretraining setting.

\paragraph{\textbf{Curriculum Learning Method}}
We set three different experiment settings to see how curriculum learning influences performance. The first setting is \textit{No Curr}, which randomly selects one ratio out of ten ranging from 0.01 to 0.1 with 0.01 interval for every iteration. The second and third settings are \textit{Continuous} and \textit{Discrete} explained in \ref{curriculum}. For all tasks except for QQP, the discrete setting performs best. Continuous method goes over simple examples too fast, leading to confusion rather than fast convergence \cite{CurrPower}.

\begin{figure}[!ht]
\centering\includegraphics[width=0.4\textwidth]{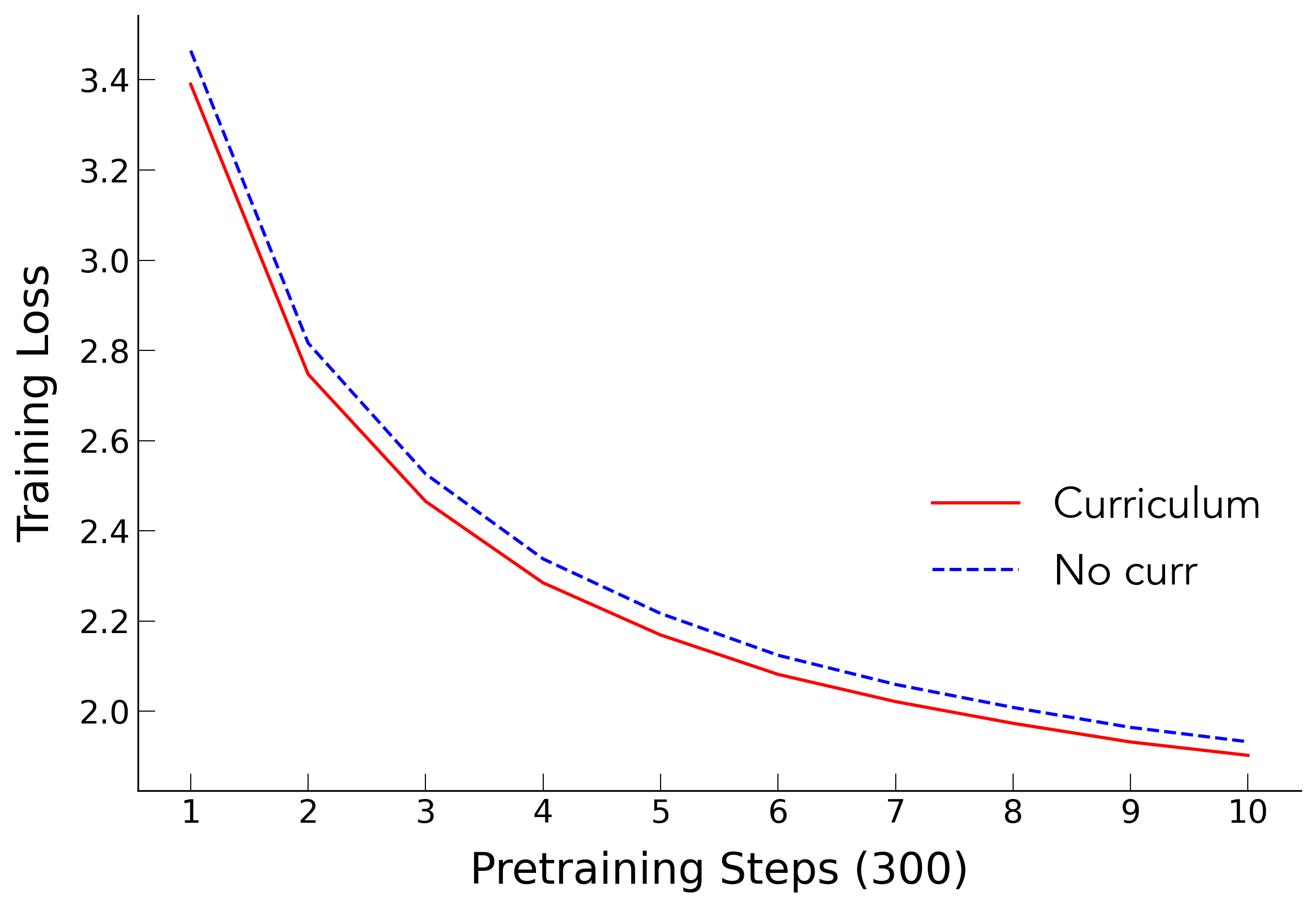}
\caption{Average batch training loss of first 3,000 steps with and without curriculum learning. }
\label{fig:curriculum}
\vspace{-5mm}
\end{figure}

Looking at the effect of curriculum learning, it is effective for most of the tasks, especially for QNLI and SST. It facilitates faster convergence to the pretraining objective as shown in Figure \ref{fig:curriculum}, leading to better performance on downstream tasks. Surprisingly, the method without curriculum learning results in the highest performance on CoLA. Due to catastrophic forgetting suggested in \cite{NLPCurr}, EffectiveCL is likely to be robust to larger noise at the end of training. Therefore, because random shuffling can better differentiate small noise, it is suitable for noise-sensitive tasks. 

\subsection{Efficiency and Applicability}
\begin{figure}[!ht]
\centering
\includegraphics[width=0.4\textwidth]{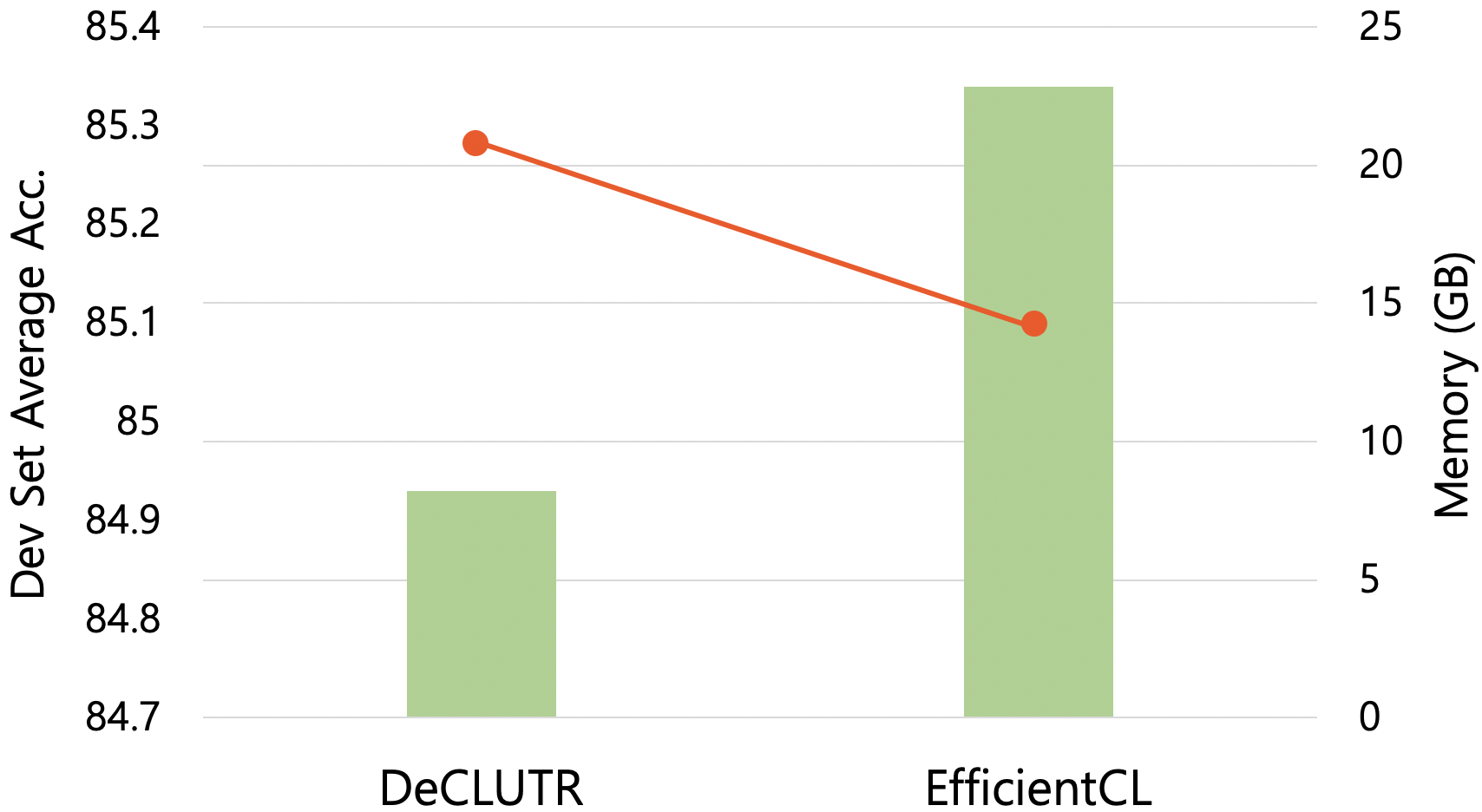}
\caption{Efficiency of our model compared with DeCLUTR.}
\label{fig:efficiency}
\vspace{-5mm}
\end{figure}

Another advantage of EfficientCL is its memory efficiency. We compare with DeCLUTR model since both train with continual pretraining. Figure \ref{fig:efficiency} shows that with 70\% of memory, our model performs better on GLUE. By applying data augmentation on the anchor for contrastive learning instead of sampling neighboring positives, our model needs only one anchor to sample, leading to reduced computational costs but better performance.  

Additionally, for our model, pretraining is possible with documents having more than 512 tokens. This is significant for applicability since DeCLUTR needs at least 2048 tokens for each document to sample the anchors and the positive spans. From the OpenWebText corpus, documents with more than 512 tokens result in 4,126,936 documents, more than 8 times compared to the current setting. Although we used the same pretraining data as DeCLUTR for fair comparison, using more documents would lead to better performance. 

%% file: conclusion.tex
In this paper, we propose EfficientCL, a pretrained model with efficient contrastive learning method utilizing data augmentation and curriculum learning. Our data augmentation process is divided into \textit{cutoff} and \textit{PCA jittering}. Rather than using one, combining two augmentation methods significantly boosts the performance. Additionally, by incrementing the augmentation level for each difficulty step, the model achieves faster convergence, resulting in better performance on sentence-level tasks.
Our method is also memory efficient and applicable for a wide range of corpora at pretraining stage.

%% file: appendix.tex
\begin{table}[!t]
\small
\centering
\begin{tabular}{@{}l|ccc}
\toprule
Model & Roberta & Continual Roberta & EfficentCL \\ \midrule
CoLA & \textbf{59.38} & 59.07 & 59.05\\
MNLI & 87.76 & 87.65 & \textbf{87.80}\\
MRPC & 90.07 & 90.14 & \textbf{90.68}\\
QNLI & 92.44 & 92.60 & \textbf{92.81}\\
QQP & 89.44 & 89.47 & \textbf{89.51}\\
RTE & 76.90 & 76.90 & \textbf{77.62}\\
SST & 94.15 & 93.92 & \textbf{94.61}\\
STSB & 90.55 & 90.45 & \textbf{90.61}\\
\midrule
Avg. & 85.09 & 85.03 & \textbf{85.34}\\
\bottomrule
\end{tabular}
\caption{Evaluation result on GLUE development set. All performance is median of five runs with random seeds.}
\label{table:continual_roberta}
\vspace{-5mm}
\end{table}

\subsection{Variants of Baseline Models}
\label{sec:variant_baseline}
\hspace{4mm}\textbf{Continually Pretrained RoBERTa-Base}

For a fair comparison of pretrained RoBERTa-base, we also continually pretrain this baseline on OpenWebText. All the settings such as sampling procedure and batch size are the same as EfficientCL except that none of data augmentation, curriculum learning, or contrastive learning is used. Only a naive MLM objective is used for continual pretraining.
Table \ref{table:continual_roberta} shows that EfficientCL outperforms continually pretrained RoBERTa for all tasks except for CoLA, which shows comparable results. For many tasks (CoLA, MNLI, SST, STSB), continually pretrained RoBERTa even underperforms pretrained RoBERTa. This shows that straightforward continual pretraining is ineffective because training hyperparameters such as learning rate scheduling are completely changed from the pretraining stage. In contrast, our EfficientCL shows robust performance, although the training objective has changed for continual pretraining setting.

\hspace{4mm}\textbf{Naive Curriculum Learning Method}

Many traditional ways of applying curriculum learning on natural language use sentence length, word rarity, or additional teacher model to evaluate the difficulty for each sentence \cite{NLPCurr}. However, these naive methods are infeasible at the continual learning stage. First of all, using multiple teacher models is memory inefficient since many language models are needed. Also, sorting sentences in respect to sentence length or word rarity is a huge overhead when the corpus is large. Compared to these naive methods, our curriculum learning, which augments the noise level of data augmentation for contrastive learning, is efficient.

\subsection{Hyperparameter Settings}
\label{sec:hyperparam}
\hspace{4mm}\textbf{Number of Anchor}

Different from DeCLUTR \cite{DeCLUTR}, we sample one anchor per document instead of two. Multiple anchors improve the performance by \textit{hard negative mining} because hard negative mining is effective for contrastive learning \cite{HardNegMix, CLHard}. Sampling multiple anchors from the same document would lead to mining negatives that are similar, resulting in a harder task for the model to learn. However, we empirically found out that the number of epoch for the model to converge at pretraining stage is insufficient when training with 2 anchors with data augmentation. This is a tradeoff between training efficiency and model performance. We expect that more training epochs with multiple anchors per document would lead to better performance.
 
\textbf{Data Augmentation Ratio}
 
For cutoff method, \citet{Cutoff} shows that there exists a sweet point for good performance of span cutoff, which is cropping ratio from 0 to 0.1. For PCA jittering method, the original paper set the standard deviation as 0.1. We found out that this value is also appropriate for large dimensions of hidden states. 

\textbf{Curriculum Learning Step}

\citet{NLPCurr} suggests that the difficulty step of 10 is appropriate for curriculum learning in natural language domain. 
 
 \textbf{Experiments}
 
For continual pretraining setting, we apply contrastive learning with a minibatch size of 4 and a temperature of \(\tau\) = 0.05. The model is trained for 1 epoch using AdamW optimizer with a learning rate 5e-05 and 0.1 weight decay. Slanted triangular LR scheduler is used for the scheduler, and gradients are scaled to 1.0 norm. 

For finetuning evaluation setting, it is done with a minibatch size of 16 and optimized using Adam optimizer with learning rate of 1e-05. The number of epochs is set as 3 epochs for small datasets (CoLA, STSB, MRPC, RTE) and 10 epochs for the rest (MNLI, QQP, SST, QNLI). \citet{FewSample} showed that conventional 3 epoch finetuning is suboptimal for small datasets due to instability. We also empirically found out that training for longer epochs significantly improves the performance for small datasets.